\documentclass[acmsmall]{acmart}
\usepackage{multirow}
\usepackage{verbatim}
\AtBeginDocument{%
  \providecommand\BibTeX{{%
    \normalfont B\kern-0.5em{\scshape i\kern-0.25em b}\kern-0.8em\TeX}}}

\setcopyright{acmcopyright}
\copyrightyear{2022}
\acmYear{2022}
\acmDOI{XXXXXXX.XXXXXXX}

\acmJournal{JACM}
\acmVolume{37}
\acmNumber{4}
\acmArticle{111}
\acmMonth{8}

\begin{document}
\title{Scale-Semantic Joint Decoupling Network for Image-text Retrieval in Remote Sensing}


\author{Chengyu Zheng}
\email{zhengchengyu@stu.ouc.edu.cn}

\author{Ning Song}
\email{songning@stu.ouc.edu.cn}
\author{Ruoyu Zhang}
\email{zhangruoyu@stu.ouc.edu.cn}
\author{Lei Huang}
\email{huangl@ouc.edu.cn}
\author{Zhiqiang Wei}
\email{weizhiqiang@ouc.edu.cn}
\author{Jie Nie}
\email{niejie@ouc.edu.cn}
\makeatletter
\let\@authorsaddresses\@empty
\makeatother
\thanks{Authors' address:  Chengyu Zheng, zhengchengyu@stu.ouc.edu.cn; Ning Song, songning@stu.ouc.edu.cn; Ruoyu Zhang, zhangruoyu@stu.ouc.edu.cn; Lei Huang, huangl@ouc.edu.cn; Jie Nie (corresponding author), niejie@ouc.edu.cn, College of Information Science and Engineering, Ocean University of China, China.}
\affiliation{
	\institution{College
		of Information Science and Engineering, Ocean University of China}
	\country{China}
}

\renewcommand{\shortauthors}{}


\begin{abstract}
Image-text retrieval in remote sensing aims to provide flexible information for data analysis and application. In recent years, state-of-the-art methods are dedicated to ``scale decoupling'' and
``semantic decoupling'' strategies to further enhance the capability of representation. However, these previous approaches focus on either the disentangling scale or semantics but ignore merging these two ideas in a union model, which extremely limits the performance of cross-modal retrieval models. To address these issues, we propose a novel Scale-Semantic Joint Decoupling Network (SSJDN) for remote sensing image-text retrieval. Specifically, we design the Bidirectional Scale Decoupling (BSD) module, which exploits Salience Feature Extraction (SFE) and Salience-Guided Suppression (SGS) units to adaptively extract potential features and suppress cumbersome features at other scales in a bidirectional pattern to yield different scale clues. Besides, we design the Label-supervised Semantic Decoupling (LSD) module by leveraging the category semantic labels as prior knowledge to supervise images and texts probing significant semantic-related information. Finally,  we design a Semantic-guided Triple Loss (STL), which adaptively generates a constant to adjust the loss function to improve the probability of matching the same semantic image and text and shorten the convergence time of the retrieval model. Our proposed SSJDN outperforms state-of-the-art approaches in numerical experiments conducted on four benchmark remote sensing datasets.

\end{abstract}

\ccsdesc[500]{Computer systems organization~Embedded systems}
\ccsdesc[300]{Computer systems organization~Redundancy}
\ccsdesc{Computer systems organization~Robotics}
\ccsdesc[100]{Networks~Network reliability}

\keywords{Remote sensing, scale-semantic joint decoupling, image-text retrieval }

\maketitle

\section{Introduction}
With the rapid development of satellite sensors, the quantity and quality of remote sensing (RS) data are increasing rapidly. To improve the utilization efficiency of remote sensing data, a large number of researches have been carried out on RS retrieval tasks \cite{hoxha2020toward,xiong2020deep,liu2019adversarial}. Compare with RS unimodal retrieval, the RS multimodal retrieval has more significance because it can mine richer semantic information and capture a more comprehensive representation. Besides, image and text cross-modal retrieval plays an important role in multimodal retrieval. It has the capability to explore the potential correlation between image and text data and realize the complementarity of the two modal data, which has attracted a great amount of attention from more and more researchers.
\begin{figure}
	\centering
	\includegraphics[width=0.65\linewidth]{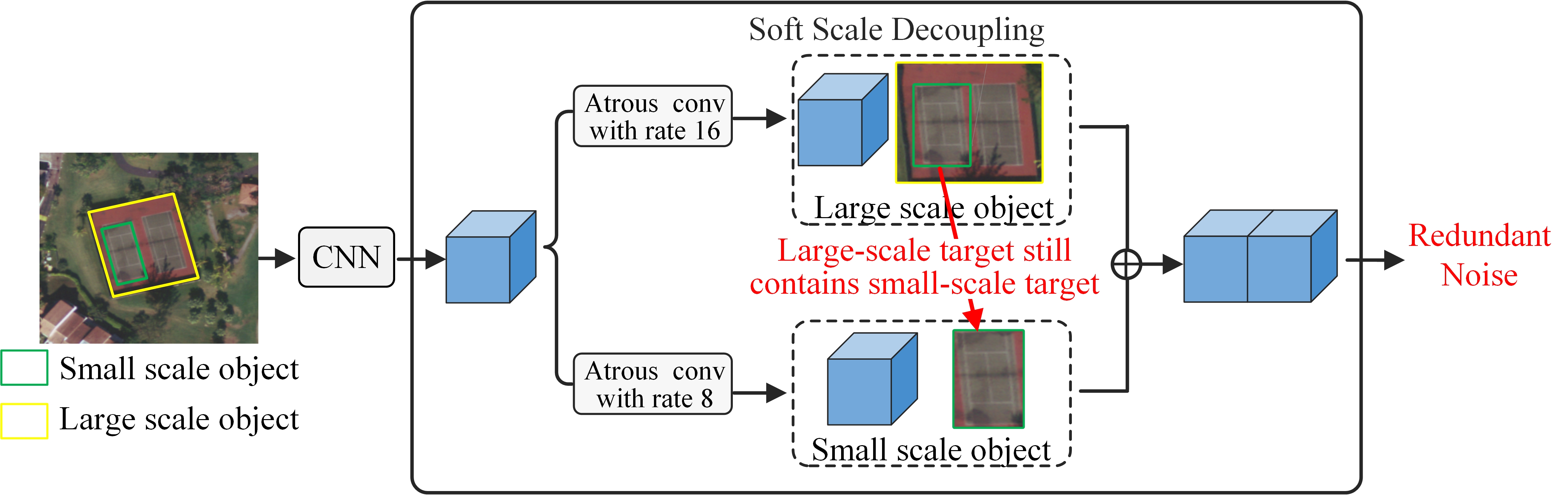}
	\caption{Here, we illustrated the ``soft scale decoupling'' and it can be seen that after scale decoupling by CNN and atrous convolution, the large-scale object tennis court in the yellow box still contains the small-scale object tennis court in the green box. This situation will eventually lead to some severe troubles such as in the subsequent fusion process, the repetitive regions are continuously accumulated, thereby bringing a great amount of redundant noise and resulting inferior utilization of multi-scale contents.}
	\label{fig1}
\end{figure}

The RS image-text cross-modal retrieval framework mainly includes three stages, namely feature extraction, feature index, and feature similarity measuring. The goal of the feature extraction stage is to mine more abstract information in data by leveraging deep learning methods, such as that Napoletano et al. \cite{napoletano2018visual} exploited the powerful feature representation capability of convolutional neural networks (CNN) to extract high-dimensional feature representations from remote sensing images and texts. Recently, state-of-the-art methods are dedicated to ``scale decoupling'' and ``semantic decoupling'' strategies to further enhance the capability of representation. The ``scale decoupling'' means that objects of different scales are modeled separately, thus more comprehensively considering the different scales of various objects, while the ``semantic decoupling'' hold the capability of entangling the semantic-mixing features, which can not only remove background noise but also enhance the inner-semantic consistency and intra-semantic discriminability. Due to the complexity and low resolution of RS images, the ``scale decoupling'' is often used on images rather than text.
For example, aiming at the multi-scale characteristics of RS targets, Cheng et al. \cite{cheng2021deep} designed an Asymmetric Multimodal Feature Matching Network (AMFMN) to simultaneously mine small-scale and large-scale feature representations for RS images. With regard to ``semantic decoupling'', Lee et al. \cite{lee2018stacked} designed the Stacked Cross Attention Network (SCAN), which first calculates a sentence/word representation for each region in the image based on similarity, and then compares it with the global sentence to achieve semantic decoupling of regions. 
The feature indexing stage is able to embed high-dimensional features into low-dimensional spaces to achieve efficient search and storage. Zou et al. \cite{zou2018novel} represented a rotation-invariant hash network to speed up the retrieval process based on the rotation invariance of RS targets.
The feature similarity measuring determines the matching degree of the two data by computing the similarity between the query data and the database, and related studies \cite{liu2020similarity,liu2020high} proposed various loss functions based on similarity measures to train end-to-end RS cross-modal retrieval networks.

Despite the significance and value of the ``scale decoupling'' and ``semantic decoupling'' methods in the feature extraction stage, these previous ``semantic decoupling'' approaches focus on either the disentangling scale or semantic but ignore merging these two ideas in a union model, which extremely limits the performance of cross-modal retrieval models. Besides, the mentioned ``semantic decoupling'' approach is only considered as a kind of ``soft scale decoupling'' where the large-scale targets still contain the small-scale targets after the decoupling operation. Here, we illustrated the ``soft scale decoupling'' in Figure 1 and it can be seen that after scale decoupling by CNN and atrous convolution, the large-scale object tennis court in the yellow box still contains the small-scale object tennis court in the green box. This situation will eventually lead to some severe troubles such as in the subsequent fusion process, the repetitive regions are continuously accumulated, thereby bringing a great amount of redundant noise and resulting inferior utilization of multi-scale contents.

To tackle these downsides, we propose an imploded framework, called Scale-Semantic Joint Decoupling Network (SSJDN) to perform both ``scale decoupling'' and ``semantic decoupling'' for RS image-text retrieval. The SSJDN follows the classical retrieval method, which first extracts the features of images and texts, and measures the similarity between the obtained features.
Unlike preceding methods, in the image feature extraction stage, we propose a Bidirectional Scale Decoupling (BSD) module, which exploits Salience Feature Extraction (SFE) and Salience-Guided Suppression (SGS) units to adaptively extract potential features and suppress cumbersome features at other scales in a bidirectional pattern to yield different scale clues. Besides, we design the Label-supervised Semantic Decoupling (LSD) module, which first leverages the category semantic labels as prior knowledge to generate category features, and then applies these category features to image features by multiple operations to decoupling semantics and probing significant semantic-related information. What's more, in the feature similarity measuring stage, we design a Semantic-guided  Triple Loss (STL) by performing category matching on two modal features and outputting a constant to adjust the loss function, which can improve the probability of matching the same semantic image and text and shorten the convergence time of the retrieval model. 
Experiments conducted on three RS image-text retrieval datasets demonstrate the effectiveness of the proposed architecture.

The contributions of this paper are as follows:

\begin{itemize}
	\item We propose an imploded framework SSJDN to perform both ``scale decoupling'' and ``semantic decoupling'' for RS image-text retrieval. By applying such a joint decoupling strategy, the feature representation capability can be crucially improved.

	\item We propose a BSD module to adaptively extract potential features and suppress cumbersome features at other scales in a bidirectional pattern to exploit distinct clues. 
	Besides, we design an LSD module, where the category semantic labels are leveraged to supervise the network decoupling image semantics and probing significant semantic-related information. Finally, we present an STL module by generating a constant to improve the probability of matching the same semantic image and text and shorten the convergence time of the retrieval model.
	
	\item We validate the effectiveness of the proposed method on four public RS image-text retrieval datasets to demonstrate the superiority of our approach.
\end{itemize}


\section{Related Work}
\subsection{Image-text Retrieval of IS}
The RS image-text retrieval is defined as, given a query image (or text), the retrieval network needs to find the corresponding labeled text (or image) from the database according to the measure of similarity. 
The automatic image caption is the earliest derived cross-modal retrieval method and has gradually developed into the mainstream of cross-modal retrieval \cite{zhang2017natural,zhang2019multi,wang2019semantic}. Shi et al. \cite{shi2017can} proposed an RS image captioning framework by leveraging fully convolutional networks to express image element attributes and interaction. To provide data assistance for the verification of the caption-based retrieval methods, Lu et al. \cite{lu2017exploring} published the largest dataset, termed RSICD. Hoxha et al.\cite{hoxha2020new} designed the CNN-RNN framework, which integrates with beam-search to output multiple captions for the target image and takes the caption with the highest similarity as the matching result. Aiming at the duplicate data of RSICD, Li et al. \cite{li2020truncation} explored the overfitting problem in RS image captioning caused by cross-entropy (CE) loss and accordingly propose a new truncated cross-entropy (TCE) loss.
Lu et al. \cite{lu2019sound} performed an active attention framework for more specific caption generation according to the interest of the observer. Although significant progress has been made in caption-based RS image retrieval, there are still some drawbacks, for example, the generated sentence is always coarse which hinders the widespread application of RS multimodal data. Recently, a small amount of RS image-text retrieval research are sprouting. In 2019, Abdullah et al. represented the Deep Bidirectional Triplet Network (DBTN) \cite{abdullah2020textrs} to solve the RS image-text retrieval problem for the first time and published a new dataset, called TextRS. Aiming at the multi-scale and complex semantic characteristics of remote sensing data, the state-of-the-art retrieval methods are dedicated to two aspects, namely “scale decoupling” and “semantic decoupling”. We will elaborate on these two strategies as follows.

\subsection{Scale Decoupling and Semantic Decoupling}
Generally, both ``scale decoupling'' and ``semantic decoupling'' are applied in the feature extraction stage. In recent years, ``scale decoupling'' methods have proved to have obvious advantages in the field of vision due to their capability to entirely explore the various objects of images, and have been widely adopted by related research.
To restore multi-scale spatial information, the ``encoder-decoder'' structure, called ``U-Net'' \cite{ronneberger2015u} was proposed. This framework concatenates the obtained low-level features with high-level features by skip connection. Szegedy et al. \cite{szegedy2015going} proposed a multi-scale model based on four parallel inception modules, which used convolution kernels of different scales for feature extraction and pooling.
The Atrous Spatial Pyramid Pooling (ASPP) \cite{chen2018encoder} focuses on the receptive field to fuse multi-scale contextual information with different dilated rates, increasing the discriminative power of the resulting feature representation. 
Undoubtedly, some studies also utilized scale decoupling methods to extract the various object of RS images in the process of RS retrieval. For example, Ye et al. \cite{ye2017multiple} proposed a flexible multiple-feature hashing learning framework for RS image retrieval by taking multi-scale complementary features. Sukhia et al. \cite{sukhia2020content} discussed a content-based RS image retrieval technique using multi-scale, patch-based local ternary pattern and fisher vector encoding. The AMFMN \cite{yuan2021exploring} also enhanced the feature representation by integrating multi-scale features into low-level and high-level features. 
With regard to ``semantic decoupling'', Lee et al. [ 10] designed the Stacked Cross Attention Network (SCAN), which first calculates a sentence/word representation for each region in the image based on similarity, and then compares it with the global sentence to achieve semantic decoupling of regions. To refine the correspondence between RS images and text, Cheng et al. introduced a Deep Semantic Alignment Network (DSAN) \cite{cheng2021deep} by utilizing a Semantic Alignment Module (SAM) to mine the underlying semantic relationship between RS images and text.

Despite the significance and value of the ``scale decoupling'' and ``semantic decoupling'' methods, there are still some problems. Firstly, these previous “semantic decoupling” approaches focus on either the disentangling scale or semantic but ignore merging these two ideas in a union model, which extremely limits the performance of cross-modal retrieval models.  Besides, the mentioned “semantic decoupling'' approach is only considered as a kind of ``soft scale decoupling'' where the large-scale targets still contain the small-scale targets after the decoupling operation, and this situation will eventually lead to a great amount of redundant noise. Moreover, these ``semantic decoupling'' methods utilize attention to implicitly mining semantics, which does not consider the application of semantic tags to explicitly mine meaningful information, thus failing to achieve the best semantic decoupling performance.

\section{Approach}
\begin{figure*}
	\centering
	\includegraphics[width=1\linewidth]{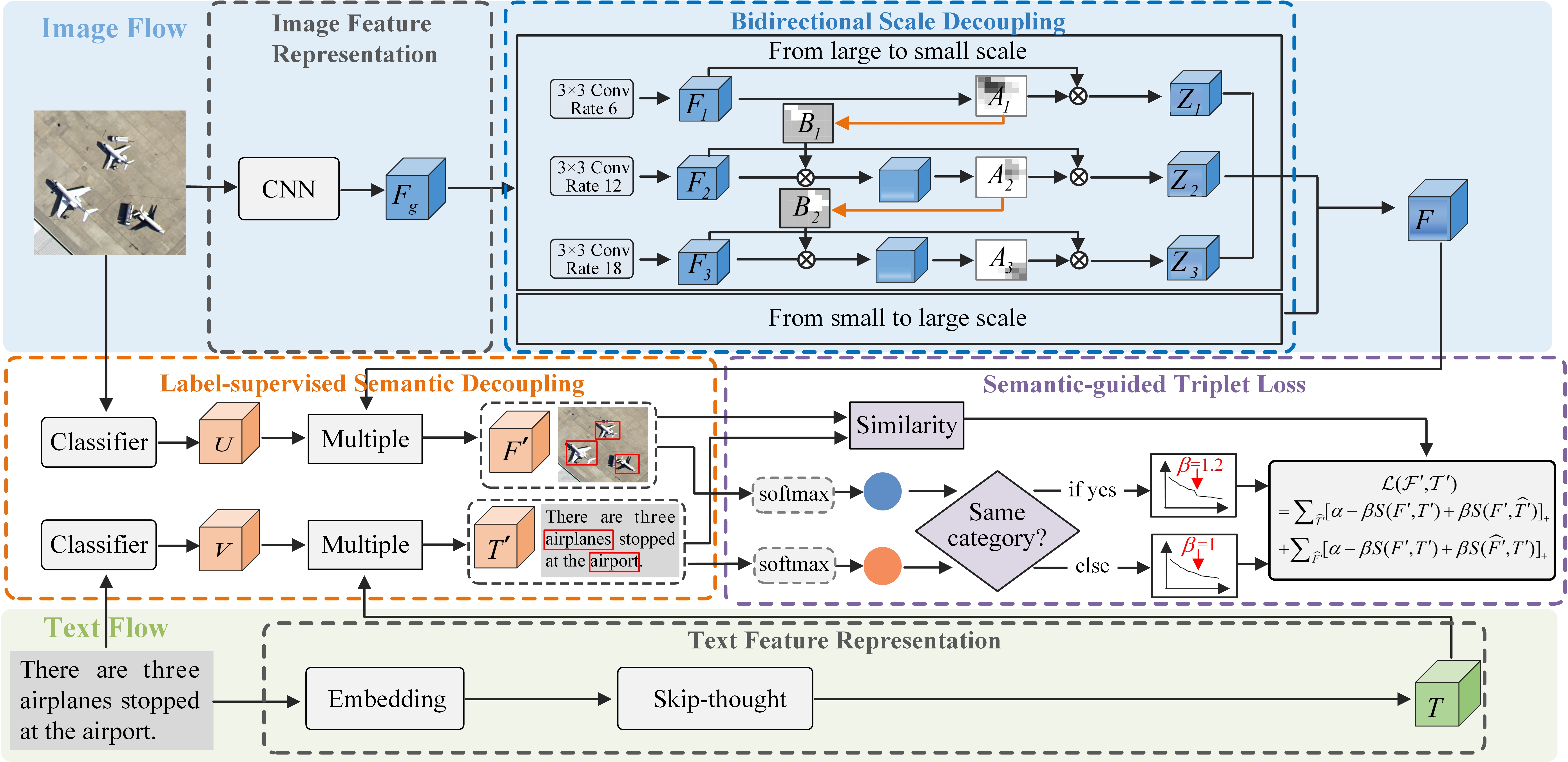}
	\caption{The retrieval architecture (SSJDN) is proposed in this paper. Especially, we represent the BSD module to adaptively filter redundant information between different scales, and the LSD module to capture more effective semantic-related semantic information. We also design a STL, which can further enhance the probability of the same class cross-modal data.}
	\label{fig2}
\end{figure*}
In this section, we introduce the main parts of our SSJDN approach. Figure 2 illustrates the framework of our approach, which includes four components. (1) Feature Representation, which could extract features of RS images and texts. (2) The Bidirectional Scale Decoupling (BSD) module, which utilizes Salience Feature Extraction (SFE) and Salience-Guided Suppression (SGS) units to adaptively extract potential features and suppress cumbersome features at other scales in a bidirectional mode to provide different clues. (3)The Label-supervised Semantic Decoupling (LSD), which applies category semantic labels to supervise the network focusing on predominant semantic-related features.  (4) The Semantic-guided Triple Loss (STL)  is designed to increase the retrieval opportunities of the same category cross-modal data.
It is worth noting that the former three parts are performed in the feature extraction stage, and the last part is in the feature similarity measuring stage. The entire retrieval process can be divided into the following four steps.

1) enter the query image (or query text) and all the samples of the dataset;

2) input the query image (or query text) and all the samples of the dataset into the trained model to obtain the visual feature (or text feature) of the query image (or query text) and the samples' feature of the dataset;

3) calculate the similarity score between the generated query visual features (or query text features) and all the samples in the dataset;

4) the samples finally are ranked according to the similarity score and returned as the search result.

Next, we will provide details of our proposed model for cross-modal image-text retrieval in remote sensing. First, we introduce the feature representation in Section III-A. Then, we describe in detail our proposed modules in Section III-C, respectively. Finally, we discuss the semantic-guided triple in Section III-D.

\subsection{Feature Representation}
We will summarize the feature representation part from two aspects: image feature representation and text feature representation.
\subsubsection{Image Feature Representation}
According to the method \cite{chen2018encoder}, for an RS image $I \in \mathbb {R}  {^{H \times W \times 3}} $, we utilze the $Resnet$ \cite{he2016deep} as feature extractor to train our model. Specifically, the global features $F_{g}$ can be formulated as: 

\begin{equation}
{F_{g}} = Resnet(I,{\theta _I}),
\end{equation}
where $\theta _I$ is used to indicate the parameters of Resnet. Subsequently, the $ASPP$ with different dilated rates are applied to generate multi-scale feature representations ${F_m}\in \mathbb {R}  {^{h \times w \times d}}$, which are defined as: 
\begin{equation}
\{ {F_m}\} _{m = 1}^3 = ASPP({F_{g}},{\theta _{{F_{g}}}}),
\end{equation}
where $m$ refers to the number of feature scales, and the dilated rates of convolution kernel corresponding to scale 1, 2, and 3 are 6, 12, and 18, respectively. $\theta _{{F_{g}}}$ is the parameters of $ASPP$.
\subsubsection{Text Feature Representation}
For texts, suppose that there are $K$ words in a sentence $S$, each word is encoded into a one-hot vector that indicates the index in the vocabulary, expressed as $ {T_S}\in \mathbb {R}  {^{K}}$. Then, the one-hot vector is converted into high-dimensional vector by the embedding matrix $W_S$, which is denoted by a formula ${T_W} = {W_S}({T_S})$. To retain temporal information in the sentence and reduce the computational cost, the high-dimensional vector is fed into a $skip-thought$ network \cite{kiros2015skip} to capture the contextual information.
\begin{equation}
T=ST(T_W,\theta _{T_W}),
\end{equation}
where $T\in \mathbb {R}  {^{d'}}$ stands for the high-level text feature and $ST$ represents the text feature extraction network, termed $skip-thought$. $\theta _{T_W}$ refers to the parameter of $skip-thought$ network.

\subsection{Bidirectional Scale Decoupling}
As mentioned above, considering the multi-scale characteristics of RS images, it is necessary to conduct multi-scale modeling for networks. However, a great amount of redundant information is continuously superimposed due to the ``soft scale decoupling''. Thus, we hope to use a kind of ``hard scale decoupling'' approach to increase the availability of multi-scale features. We will implement the ``hard scale decoupling'' in two steps, extracting potential features on the current scale firstly, and then suppressing cumbersome features at other scales. Based on the above analysis, we propose a BSD consisting of an SFE unit and an SGS unit, as shown in Figure 3. It is worth noting that the BSD is bidirectional (from small to large scale and from large to small scale), which can extract feature representations more comprehensively and reliably.

\subsubsection{Salience Feature Extraction}
The SFE unit is proposed to extract potential features on the current scale. In recent years, the Convolutional Block Attention Module (CBAM) \cite{woo2018cbam} has prominent performance in exploring the discriminative information by calculating the attention maps of features. Therefore, we prefer to use the CBAM to construct the SFE unit. Aiming at the different scale features $F_m $, we first aggregate the channel-wise information of a feature by the average pooling and max pooling operations, and generate two 2D efficient feature descriptor: $P_{m}^ {avg}(F_m)\in\mathbb{ R} ^{h\times w}$ and $P_{m}^ {avg}(F_m)\in\mathbb{ R} ^{h\times w}$. Then, the feature descriptor passes through a standard convolution layer and  sigmoid function to produce the attention map $A_m$: 
\begin{equation}
A_m=\sigma (conv(concat(P_{m}^ {avg}(F_m),P_{m}^ {max}(F_m)))),
\end{equation}
where $\sigma$ denotes the sigmoid function and $conv(\cdot )$
represents a convolution layer. $concat(\cdot, \cdot)$ means concatenation of the features. The attention maps $A_m$ are effective in highlighting informative regions of features in different scales. 
\begin{figure}
	\centering
	\includegraphics[width=0.4\linewidth]{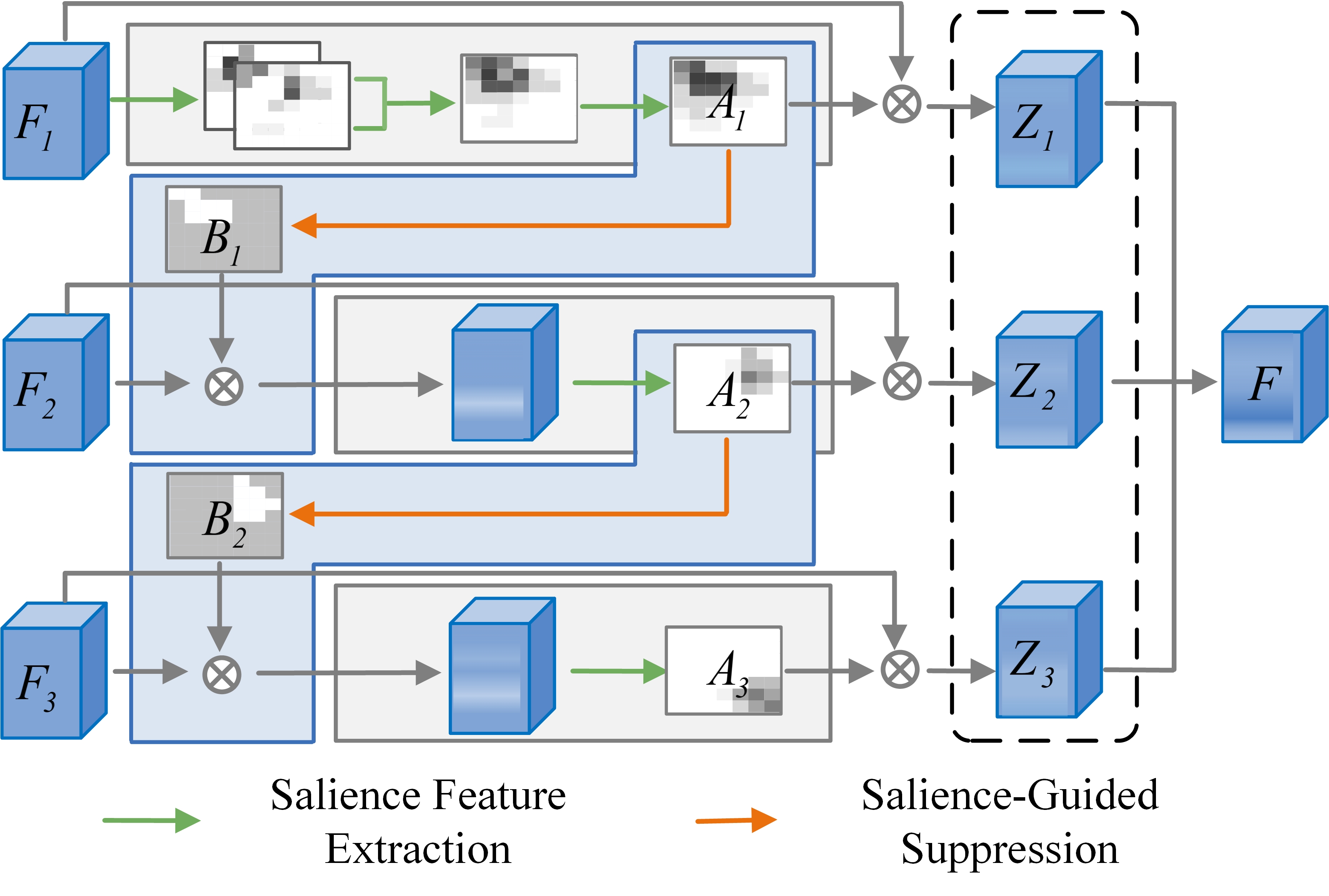}
	\caption{A illustration of an unidirectional (from small to large scales) scale decoupling (USD) module. The attention map $A_m$ is used to extract the salient features, and the suppression mask $B_m$ is prepared for suppressing the salience features at other scales.}
	\label{fig3}
\end{figure}
\subsubsection{Salience-Guided Suppression}
After completing salience feature extraction, the SGS unit is created to perform suppression operations on salience features at the current scale, benefiting the other scales features to explore discriminative cues.
Specifically, due to the attention map, $A_m$ represents the salience region of the feature $F_m$,
the suppression mask $B_m$ makes full use of $A_m$ to restrain the salience information at the scale $m$:
\begin{equation}
B_m=\mathcal{B} (A_m),
\end{equation}
where $\mathcal{B}$ is a binary mask, which takes the values of the most salience $A_m$ to 0 and others to 1. The suppression masks relieve the coverage effect of $A_m$ on other scales and make different information stand out.

Subsequently, based on SFE and SGS, we build the BSD from two directions. For the sake of simplicity, Figure 3 only shows a unidirectional (from small to large scales) scale decoupling module. First, the attention map $A_m$ and the salience mask $B_m$ are calculated, where the $A_m$ is used to extract the salience features, and the $B_m$ is prepared for suppressing the salience features at other scales. Then, these two maps are applied in the process of generating decoupled feature $z_m$ by way of stepwise suppression. Besides, we also add residual connections \cite{he2016deep} to obtain more effective information, and the scale decoupled feature $Z_m$ is computed as:

\begin{equation}
\begin{cases}
Z_m=A_mF_m+F_m & \text{ if } m=1\\
Z_m=A_m(B_{m-1}F_m)+F_m & \text{ if }m=2,3
\end{cases}
\end{equation}

Similarly, the scale decoupled features $\hat{Z_m}$ in the way of large to small scale is calculated by the following formula:
\begin{equation}
\begin{cases}
\hat{Z_m}=A_mF_m+F_m & \text{ if } m=3\\
\hat{Z_m}=A_m(B_{m+1}F_m)+F_m & \text{ if }m=1,2
\end{cases}
\end{equation}


Finally, the scale decoupled features $Z_m$ and $\hat{Z_m}$ are fused by concatenation operations. 
\begin{equation}
F=concat(Z_1,Z_2,Z_3,\hat{Z_1},\hat{Z_2},\hat{Z_3}),
\end{equation}
where $F\in\mathbb{ R} ^{h\times w\times d'} $ represents the scale decoupled features of images. It can be seen that the $F$ alleviate the redundant information between multi-scale features and enhance the robustness of image representation.
\subsection{ Label-supervised Semantic Decoupling}
As discussed above, to moderate the effect of the meaningless semantic features caused by the implicit exploration of attention mechanism, we propose the LSD module to mine effective semantic features. The category semantic labels are utilized to train image and text classification networks to produce category features, which are then applied to two aspects. First, category features are multiplied with the decoupled image features and text features to guide the retrieval network probing significant and reliable semantic-related information. Second, category matching is performed on the category features, determining whether the images and texts belong to the same category for improving the retrieval probability of the same class cross-modal data, which will be detailed in the next subsection.
\subsubsection{Category Features Extraction}
We first implement image and text classification separately on each dataset, and fuse the pre-trained classification network into the retrieval network. Specifically, the image classification network with the same structure as $Resnet$ under the supervision of the image category semantic labels $\hat{Y_I}$ with cross-entropy loss to obtain the high-level category feature representation. Besides, to better fine-tune the text feature representation, we utilize the $BERT$ as a text classifier to categorize the sentence $S$ based on text category semantic labels $\hat{Y_S}$.
The category features $U$ and $V$ of image and text are calculated as follows:
\begin{equation}
\begin{aligned}
&U=Resnet(I,{\theta _{I}}' )\\&V=BERT(S,{\theta _{S}}),
\end{aligned}
\end{equation}
where ${\theta _{I}}'$ and $\theta _{S}$ are the parameters of $Resnet$ and $BERT$, respectively. 
\subsubsection{Category Features Fusion}

After category features extraction, the generated category features are leveraged to guide the decoupled image feature and text feature to mine significant and reliable features. Specifically, we take full advantage of multiplication, which can achieve considerable enhancement of related features in the process of feature combination to compute the final semantic decoupling image and text features $F'$ and $T'$:

\begin{equation}
\begin{aligned}
&F'=multiple(F,U)\\
&T'=multiple(T,V),
\end{aligned}
\end{equation}
where $multiple(\cdot,\cdot)$ denote multiplication, The ultimate feature $F'$ and $T'$
not only capture discriminable multi-scale semantic information, but also highlight the semantic-related reliable knowledge, thereby increasing the accuracy of retrieval network.

\subsection{Semantic-guided Triple Loss}
To implement multi-modal feature alignment, triplet loss is widely used and has become one of the mainstream loss functions in the field of multi-modal feature matching tasks. The purpose of triple loss is to increase the distance between a sample and the corresponding negative sample while reducing the distance between the sample and its positive sample as much as possible. \cite{faghri2017vse++} proposes a bidirectional triple loss for text-image matching:

\begin{equation}
\begin{aligned}
\mathcal{L(F',T')}=&\textstyle\sum_{\hat{T'}}[\alpha-S(F',T')+ S(F',\hat{T'} )]_+\\+&\textstyle\sum_{\hat{F'}}[\alpha-S(F',T')+ S(\hat{F'} ,T' )]_+ ,
\end{aligned}
\end{equation}
where $\alpha$ refers to the margin and $[\cdot ]_+ = max(\cdot , 0)$. $S(F',T')$ represent the similarity of image and text. The first sum is processed for all negative sentences $T'$ of a given image $F'$, and the second sum considers all negative images $F'$ of a given a sentence $T'$. To save the calculation cost, the loss is usually computed in each batch rather than in all training sets.

However, the data of the same category should be easier retrieved to improve the accuracy of image-text matching. Thus, we perform category matching based on the category features (as mentioned above) to automatically determine whether the input text and image are of the same category. Firstly, the category features are converted into semantic categories $Q_U$ and $Q_V$ (a total of c categories in the retrieval dataset) of image and text through softmax. Then, we define a constant value $\varepsilon$ to adjust the loss to make it more sensitive to the same category of different modal data. The constant value is expressed as:

\begin{equation}
\beta=\begin{cases}
\varepsilon & \text{ if } Q_U=Q_V\\
1  & \text{ others }
\end{cases}
\end{equation}

On the basis of the constant value $\varepsilon$, we design a semantic-guided triple loss detailed as follows:
\begin{equation}
\begin{aligned}
\mathcal{L(F',T')}=&\textstyle\sum_{\hat{T'}}[\alpha-\beta S(F',T')+ \beta S(F',\hat{T'} )]_+\\+&\textstyle\sum_{\hat{F'}}[\alpha-\beta S(F',T')+ \beta S(\hat{F'} ,T' )]_+
\end{aligned}
\end{equation}

We take the image to text retrieval as an example to implement analysis on semantic-guided triple loss. Since $T'$ is the positive sample corresponding to and $F'$, $F'$ and $T'$ belong to the same category, that is to say, $Q_U$ is equal to $Q_V$, and the value of $\beta$ is $\varepsilon$. We restrict $\varepsilon$ to be a value larger than 1, that will be subjected to ablation experiments to evaluate the specific value, to improve the retrieval accuracy of positive samples. Moreover, if $F'$ and $\hat{T'}$ are the same class, similarly, the value of $\beta$ is also $\varepsilon$, but when $F'$ and $\hat{T'}$ are different classes, $Q_U$ is unequal to $Q_V$, and the value of $\beta$ is still 1, which will not increase the matching rate of negative samples of different classes.
In addition, we introduce classification loss here for better understanding of the LSD module. The classification network of image (Resnet) and text (BERT) both apply cross-entropy loss to  generate category features under the supervision of image and text semantic labels $\hat{Y_I}$ and $\hat{Y_S}$.

\begin{equation}
\begin{aligned}
&\mathcal{L}_{I}(Y_I,\hat{Y_I}) =\frac{1}{N} \sum_{i=1}^{N} \left ( -Y_I^ilog(\hat{Y}_I^i)-(1-Y_I)^ilog(1-\hat{Y}_I^i)\right )   \\
&\mathcal{L}_{S}(Y_S,\hat{Y_S}) =\frac{1}{N} \sum_{i=1}^{N} \left ( -Y_S^ilog(\hat{Y}_S^i)-(1-Y_S)^ilog(1-\hat{Y}_S^i)\right )   \label{eq},
\end{aligned}
\end{equation}
where $Y_I$ and $Y_S$ are the classification prediction results of RS images and text. $N$ refers to the number of training samples. 
So far, we have introduced our approach, which not only considers the decoupling of multi-scale features but also makes full use of category semantic labels to extract meaningful feature representations.

\section{Experiments}
To assess the effectiveness of the proposed method, some
experiments are conducted. In this section, we first give a
short description of the datasets, implementation details, metrics, and baselines. After that, we carry out
experiments involving the proposed approach through ablation studies and comparisons with state-of-the-art methods. 

\begin{table}[]
	\centering
	\caption{Comparisons of image-text retrieval experiments on UCM, RSITMD, RSICD and Sydney datasets.}
\begin{tabular}{l|ccccccc}
\hline
{\color[HTML]{000000} }                         & \multicolumn{7}{c}{{\color[HTML]{000000} UCM dataset}}                                                   \\ \hline
\multicolumn{1}{c|}{}                           & \multicolumn{3}{c|}{Image-to-Text}      & \multicolumn{3}{c|}{Text-to-Image}      &                      \\
\multicolumn{1}{c|}{\multirow{-2}{*}{Approach}} & R@1  & R@5  & \multicolumn{1}{c|}{R@10} & R@1  & R@5  & \multicolumn{1}{c|}{R@10} & \multirow{-2}{*}{mR} \\ \hline
SCAN t2i                                        & 13.9 & 45.8 & \multicolumn{1}{c|}{68.6} & 13.1 & 50.7 & \multicolumn{1}{c|}{78.2} & 45.0                 \\
SCAN i2t                                        & 12.8 & 47.0 & \multicolumn{1}{c|}{69.1} & 12.6 & 46.9 & \multicolumn{1}{c|}{72.7} & 43.5                 \\
CAMP-triplet                                    & 11.2 & 44.3 & \multicolumn{1}{c|}{65.7} & 9.9  & 46.1 & \multicolumn{1}{c|}{77.3} & 42.4                 \\
CAMP-bce                                        & 15.1 & 47.2 & \multicolumn{1}{c|}{68.6} & 11.9 & 47.2 & \multicolumn{1}{c|}{77.0} & 44.5                 \\
MTFN                                            & 10.7 & 47.6 & \multicolumn{1}{c|}{64.3} & 14.7 & 52.7 & \multicolumn{1}{c|}{81.1} & 45.2                 \\
AMFMN                                           & 14.5 & 51.2 & \multicolumn{1}{c|}{67.3} & 14.5 & 51.8 & \multicolumn{1}{c|}{80.7} & 46.7                 \\ \hline
SSJDN                                           & 17.1 & 56.7 & \multicolumn{1}{c|}{77.1} & 17.9 & 60.9 & \multicolumn{1}{c|}{85.9} & 52.6                 \\ \hline
\end{tabular}
\end{table}
\begin{table}[]
\begin{tabular}{l|ccccccc}
\hline
                                               & \multicolumn{7}{c}{RSITMD dataset}                                                                      \\ \hline
\multicolumn{1}{c|}{\multirow{2}{*}{Approach}} & \multicolumn{3}{c|}{Image-to-Text}      & \multicolumn{3}{c|}{Text-to-Image}      & \multirow{2}{*}{mR} \\
\multicolumn{1}{c|}{}                          & R@1  & R@5  & \multicolumn{1}{c|}{R@10} & R@1  & R@5  & \multicolumn{1}{c|}{R@10} &                     \\ \hline
SCAN t2i                                       & 10.1 & 28.7 & \multicolumn{1}{c|}{38.5} & 10.6 & 29.5 & \multicolumn{1}{c|}{44.5} & 27.0                \\
SCAN i2t                                       & 11.3 & 25.9 & \multicolumn{1}{c|}{40.1} & 9.8  & 29.4 & \multicolumn{1}{c|}{41.6} & 26.4                \\
CAMP-triplet                                   & 11.7 & 28.6 & \multicolumn{1}{c|}{37.1} & 8.4  & 26.9 & \multicolumn{1}{c|}{45.0} & 26.3                \\
CAMP-bce                                       & 9.0  & 24.4 & \multicolumn{1}{c|}{31.2} & 5.5  & 23.7 & \multicolumn{1}{c|}{39.2} & 22.2                \\
MTFN                                           & 10.5 & 27.8 & \multicolumn{1}{c|}{35.3} & 9.6  & 31.4 & \multicolumn{1}{c|}{47.1} & 27.0                \\
AMFMN                                          & 10.1 & 26.7 & \multicolumn{1}{c|}{41.4} & 10.5 & 34.8 & \multicolumn{1}{c|}{56.9} & 30.0                \\ \hline
SSJDN                                          & 12.2 & 29.4 & \multicolumn{1}{c|}{44.2} & 10.8 & 42.2 & \multicolumn{1}{c|}{68.9} & 34.6                \\ \hline
\end{tabular}
\end{table}
\begin{table}[]
\begin{tabular}{l|ccccccc}
\hline
                                               & \multicolumn{7}{c}{RSICD dataset}                                                                     \\ \hline
\multicolumn{1}{c|}{\multirow{2}{*}{Approach}} & \multicolumn{3}{c|}{Image-to-Text}     & \multicolumn{3}{c|}{Text-to-Image}     & \multirow{2}{*}{mR} \\
\multicolumn{1}{c|}{}                          & R@1 & R@5  & \multicolumn{1}{c|}{R@10} & R@1 & R@5  & \multicolumn{1}{c|}{R@10} &                     \\ \hline
SCAN t2i                                       & 4.3 & 11.2 & \multicolumn{1}{c|}{17.6} & 4.0 & 16.6 & \multicolumn{1}{c|}{26.6} & 13.4                \\
SCAN i2t                                       & 5.9 & 12.9 & \multicolumn{1}{c|}{19.7} & 3.8 & 16.9 & \multicolumn{1}{c|}{27.1} & 14.4                \\
CAMP-triplet                                   & 5.1 & 13.0 & \multicolumn{1}{c|}{22.1} & 4.2 & 15.3 & \multicolumn{1}{c|}{27.8} & 14.6                \\
CAMP-bce                                       & 4.4 & 10.2 & \multicolumn{1}{c|}{15.7} & 2.5 & 13.4 & \multicolumn{1}{c|}{23.9} & 11.7                \\
MTFN                                           & 5.1 & 12.6 & \multicolumn{1}{c|}{19.9} & 4.9 & 17.8 & \multicolumn{1}{c|}{29.6} & 14.8                \\
AMFMN                                          & 5.5 & 14.8 & \multicolumn{1}{c|}{23.1} & 4.0 & 17.2 & \multicolumn{1}{c|}{31.3} & 16.0                \\ \hline
SSJDN                                          & 6.5 & 19.7 & \multicolumn{1}{c|}{30.1} & 4.9 & 20.2 & \multicolumn{1}{c|}{36.5} & 19.7                \\ \hline
\end{tabular}
\end{table}

\begin{table}[]
\begin{tabular}{l|ccccccc}
\hline
                                               & \multicolumn{7}{c}{Sydney dataset}                                                                      \\ \hline
\multicolumn{1}{c|}{\multirow{2}{*}{Approach}} & \multicolumn{3}{c|}{Image-to-Text}      & \multicolumn{3}{c|}{Text-to-Image}      & \multirow{2}{*}{mR} \\
\multicolumn{1}{c|}{}                          & R@1  & R@5  & \multicolumn{1}{c|}{R@10} & R@1  & R@5  & \multicolumn{1}{c|}{R@10} &                     \\ \hline
SCAN t2i                                       & 19.0 & 50.7 & \multicolumn{1}{c|}{74.1} & 17.8 & 55.9 & \multicolumn{1}{c|}{76.6} & 49.0                \\
SCAN i2t                                       & 20.4 & 54.2 & \multicolumn{1}{c|}{67.6} & 16.1 & 57.6 & \multicolumn{1}{c|}{76.0} & 48.7                \\
CAMP-triplet                                   & 22.8 & 50.5 & \multicolumn{1}{c|}{75.9} & 15.3 & 43.1 & \multicolumn{1}{c|}{70.4} & 46.3                \\
CAMP-bce                                       & 15.6 & 49.7 & \multicolumn{1}{c|}{71.3} & 11.6 & 51.3 & \multicolumn{1}{c|}{76.2} & 46.0                \\
MTFN                                           & 21.7 & 51.6 & \multicolumn{1}{c|}{69.0} & 14.1 & 56.0 & \multicolumn{1}{c|}{78.6} & 48.5                \\
AMFMN                                          & 29.6 & 55.6 & \multicolumn{1}{c|}{67.6} & 13.7 & 60.0 & \multicolumn{1}{c|}{82.7} & 51.5                \\ \hline
SSJDN                                          & 30.4 & 50.8 & \multicolumn{1}{c|}{68.1} & 20.4 & 67.5 & \multicolumn{1}{c|}{86.8} & 54.0                \\ \hline
\end{tabular}
\end{table}

\subsection{Datasets}
We perform experiments on four benchmark RS datasets for the cross-modal image-text retrieval: UCMerced-LandUse-Captions, Sydney-Captions, RSICD, and
NWPU-RESISC45-Captions.

The UCMerced-LandUse-Captions is build by \cite{qu2016deep}, based on the UCMerced-LandUse dataset \cite{yang2010bag}. It contains land use images in 21 categories, with 100 images per category. The spatial sizes of the data files are 256 $\times $ 256 pixels, and the ground sampling distance (GSD) is 0.3048 m. The research \cite{qu2016deep} exploit five different sentences to describe every image. 

The RSICD dataset is used for the remote sensing image captioning task \cite{lu2017exploring}. The spatial sizes of the data files are fixed to 224 $\times $ 224 pixels with various resolutions. The total number of remote sensing images is 10921, with five sentences of descriptions per image.

The RSITMD dataset is supplied by \cite{yuan2021exploring} for RS cross-modal image-text retrieval. 
The images in the RSITMD dataset are selected from the RSICD dataset and provide a total of 23715 captions for 4743 images. The RSITMD is more granular and diverse in captions than the RSICD dataset.

The Sydney-Captions dataset is also provided by \cite{qu2016deep}, which is based on the Sydney dataset \cite{zhang2014saliency}. The spatial sizes of the data files are 18 000 $\times $ 14 000 pixels, and the GSD is 0.5 m. Similar to the UCMerced-LandUse dataset, five different sentences were given to describe each image \cite{qu2016deep}.

In addition, we also conduct statistics on the categories of each dataset, and the calculation results are shown in Table 2. It can be seen from the table that the RSICD and RSITMD datasets have the most scene classes of 33. The Sydney dataset contains fewer scene with 7 categories.
\begin{table}[]
	\centering
	\caption{Number of categories of UCM, RSITMD, RSICD and Sydney datasets.}
	\begin{tabular}{l|l|l|l|l}
		\hline
		& UCM & RSICD & RSITMD & Sydney \\ \hline
		Number of categories & 21  & 33    & 33     & 7      \\ \hline
	\end{tabular}
\end{table}
\subsection{Implementation Details}
In this subsection, we mainly introduce the implementation details of the proposed method. The channel $d'$ of the scale decoupled feature $F$ is the same as that of the high-level text feature, which is 512. For the semantic-guided triple loss (Eq. 13), we conduct a large number of ablation experiments and take the value of $\varepsilon$ to be 1.2 and the margin threshold $\beta $ to 0.2. All the models are implemented with PyTorch and the Adam optimizer with a 0.0002 learning rate. We set the batch size of training to 100 and train the model for 70 epochs. All the experiments are conducted on a server with one Tesla V100 SXM2 16GB GPU.
\subsection{Metrics and Baselines}
We conduct two kinds of image-text matching tasks: 1) sentence retrieval, i.e., retrieving ground-truth sentences related to the query image (I2T); and 2) image retrieval, i.e., retrieving ground-truth images related to the query text (T2I). We use the Recall at $K$ ($R@K$, $K$=1,5 and 10) as the evaluation metric, which refers to the ratio of groud truth appearing in the topK results. Besides, we also compute the average $mR$ of the recall rates of $R@K$ raised by Huang et al. \cite{huang2018learning} to more reasonably evaluate the performance of the model.

As for baselines, six classic image-text retrieval networks are compared, including SCAN \cite{lee2018stacked}, CAMP  \cite{wang2019camp} and MTFN \cite{wang2019matching}. We additionally compare SSJDN with the recently proposed AMFMN \cite{cheng2021deep}, which dedicate to RS image-text retrieval.

\subsection{Performance Comparison}
In this section, we compare our method with six baselines on four datasets. The experimental results are listed in Table 1. Due to discrepancy in different datasets, the performance of the model is divergent on different datasets, and we can conduct the following analysis:
\begin{itemize}
	\item \textbf{Results on UCM}: From the first table of Table 1, we can observe that the retrieval models achieve the best performance due to the superiority of the UCM dataset. For the metric $mR$, our approach achieves $7.4$ and $5.9$ gains compared with the best baseline MTFN and AMFMN, respectively. It is shown that SSJDN improves retrieval efficiency by implementing category matching of images and texts, attaining a competitive performance.
	
	\item \textbf{Results on RSITMD}: The results on the RSITMD dataset
	are in the second table of Table 1. The capabilities of our SSJDN are
	competitive to some state-of-the-art methods, especially the
	SCAN-based and CAMP-based models due to the negligence of multi-scale information.
	Besides, compared with AMFMN, the proposed method has improved mR by $4.6$, which verifies the effectiveness of category guidance and the bidirectional scale decoupling method.
	\item \textbf{Results on RSICD}: The RSICD dataset reduces model robustness due to its blurry image data and the comparison of models on the RSICD dataset is on the third table of Table 1. No matter in the indicator of $R@1$, $R@5$ or $R@10$, our model is superior to other retrieval models, demonstrating the remarkable ability for both I2T and T2I retrieval. The SSJDN achieves an average improvement of $3.7$ as shown by mR on the best network AMFMN. The results indicate the availability and feasibility of proposed each modules.

	\item \textbf{Results on Sydney}: It can be observed on the bottom right of Table 1 that the proposed method achieves the best $R@1$ indicator, but is slightly less effective on $R@5$ and $R@10$ since both SCAN and CAMP consider the message passing between cross-modalities. Nonetheless, the overall performance of SSJDN is still prominent, with a score of 54.0 on the mR indicator.
	The performance on the Sydney dataset strongly illustrates the reliability and robustness of the AMFMN method.
	
\end{itemize}
\begin{table}[]
	\centering
	\caption{Comparison with different modules of image-text retrieval.}
	\begin{tabular}{c|ccc|ccc}
		\hline
		\multirow{2}{*}{Approach} & \multicolumn{3}{c|}{Image-to-Text} & \multicolumn{3}{c}{Text-to-Image} \\
		& R@1        & R@5        & R@10      & R@1       & R@5       & R@10      \\ \hline
		w/o all                   & 10.7       & 23.7      & 36.1      & 9.1       & 33.0      & 42.5     \\
		w/o BSD                   & 12.1       & 27.4      & 38.7      & 8.4       & 36.6      & 61.4      \\
		w/o LSD                  & 12.0       & 26.8      & 38.6      & 8.9      & 34.8      & 56.8      \\
		w/o STL                    & 12.2       & 27.3      & 42.1      & 9.4      & 38.3      & 67.2      \\ \hline
		SSJDN                     & 12.2       & 29.4      & 44.2      & 10.8      & 42.2      & 68.9     \\ \hline
	\end{tabular}
\end{table}
\begin{table}[]
	\centering
	\caption{Comparison with different attention mechanism of image-text retrieval.}
	\begin{tabular}{c|ccc|ccc}
		\hline
		\multicolumn{1}{c|}{\multirow{2}{*}{Approach}} & \multicolumn{3}{c|}{Image-to-Text} & \multicolumn{3}{c}{Text-to-Image} \\
		\multicolumn{1}{c|}{}                          & R@1        & R@5       & R@10      & R@1       & R@5       & R@10      \\ \hline
		w/o MA                                         & 12.1       & 27.4      & 38.7      & 8.4       & 36.6      & 61.4      \\
		MA                                             & 11.2       & 27.9      & 39.8      & 10.1      & 38.9      & 65.2      \\
		USD(L2S)                                      & 11.5       & 27.6      & 40.9      & 10.7      & 39.4      & 67.9      \\
		USD(S2L)                                      & 11.6       & 28.8      & 42.3      & 9.9      & 41.0      & 68.1      \\ \hline
		BSD                                           & 12.2       & 29.4      & 44.2      & 10.8      & 42.2      & 68.9      \\ \hline
	\end{tabular}
	
	\label{tab:plain}
\end{table}

\subsection{Module Analysis}
In this section, we carried out several experiments on the RSITMD data set to further analyze the effectiveness of our model. Specifically, we first explored how each component of our framework affects the image-text retrieval results. We then displayed how the value of $\varepsilon$ influences the retrieval performance. After that, the comparisons of different attention mechanisms are implemented to prove the validity of the BSD module. Finally, we conduct experiments on the different fusion methods between categorical features and image features.

\begin{figure}
	\centering
	\includegraphics[width=0.5\linewidth]{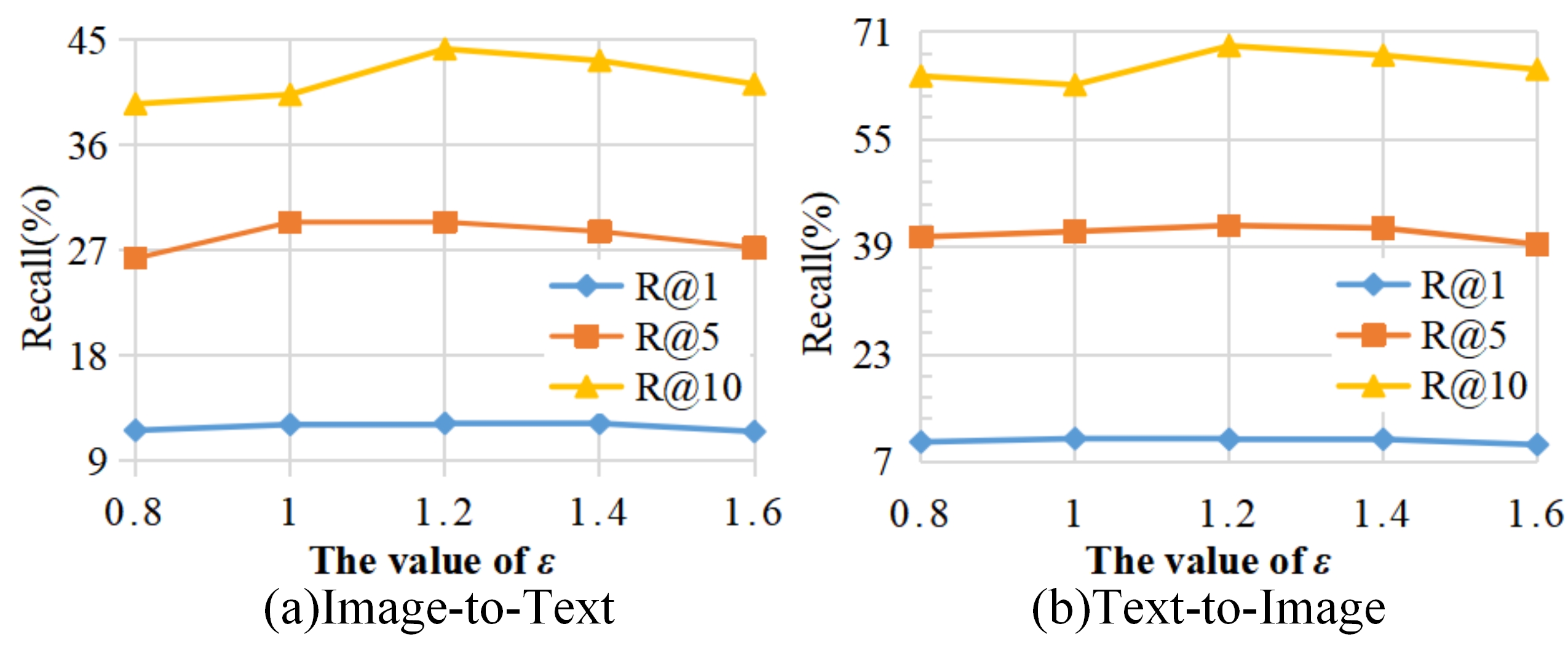}
	\caption{Performance of image-text retrieval with different scalar values of $\varepsilon$.}
	\label{fig3}
\end{figure}
\begin{figure}
	\centering
	\includegraphics[width=0.5\linewidth]{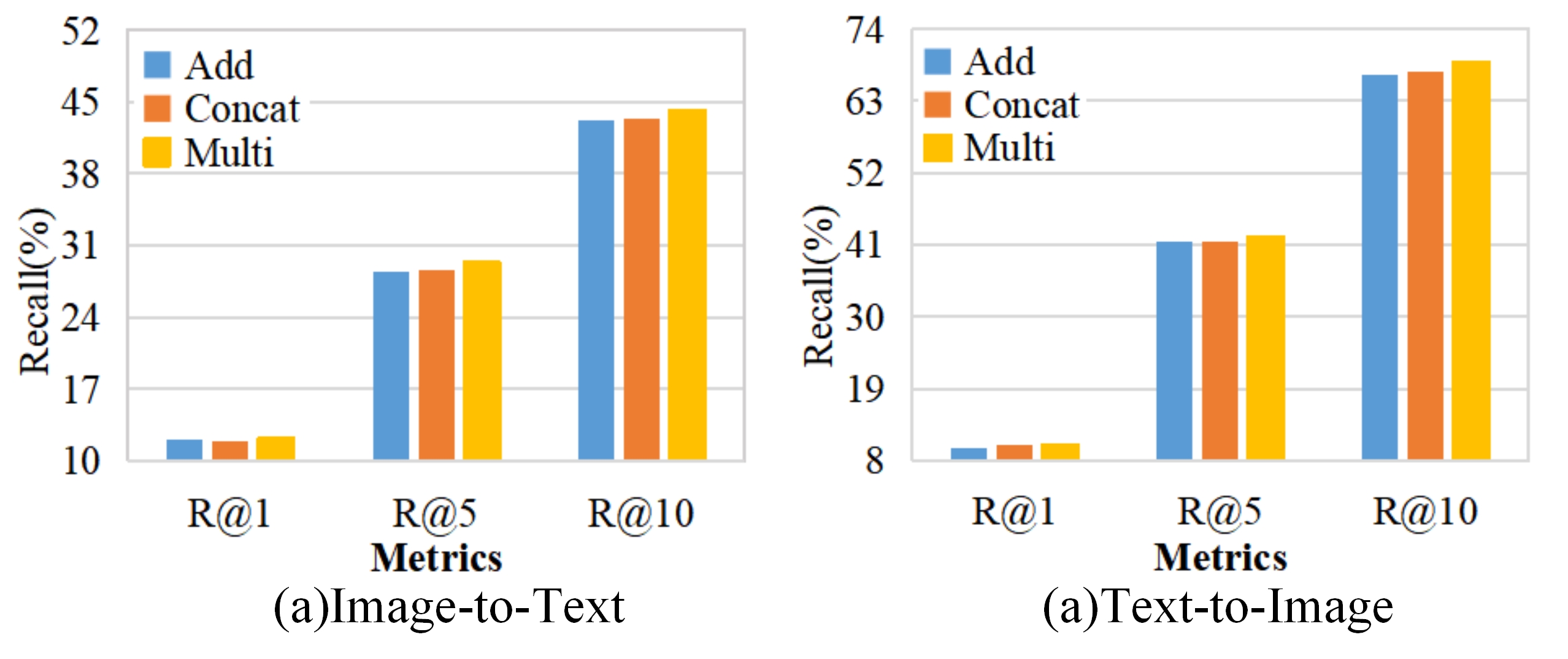}
	\caption{Performance of image-text retrieval with different fusion method.}
	\label{fig3}
\end{figure}
\subsubsection{Ablation Studies}
To verify the effectiveness of each module proposed in the paper, we conducted ablation experiments, as reported in Table 3. Specifically, we modify SSJDN with the following variants: 1)\textbf{w/o STL}, changing the semantic-guided triple loss to triplet loss; 2)\textbf{w/o LSD}, removing the label-supervised semantic decoupling module, which represent no category semantic supervision; 3)\textbf{w/o BSD}, eliminating the bidirectional scale decoupling module; 4)\textbf{w/o all}, without all components, including STL, LSD and BSD modules. 

\begin{figure*}
	\centering
	\includegraphics[width=0.95\linewidth]{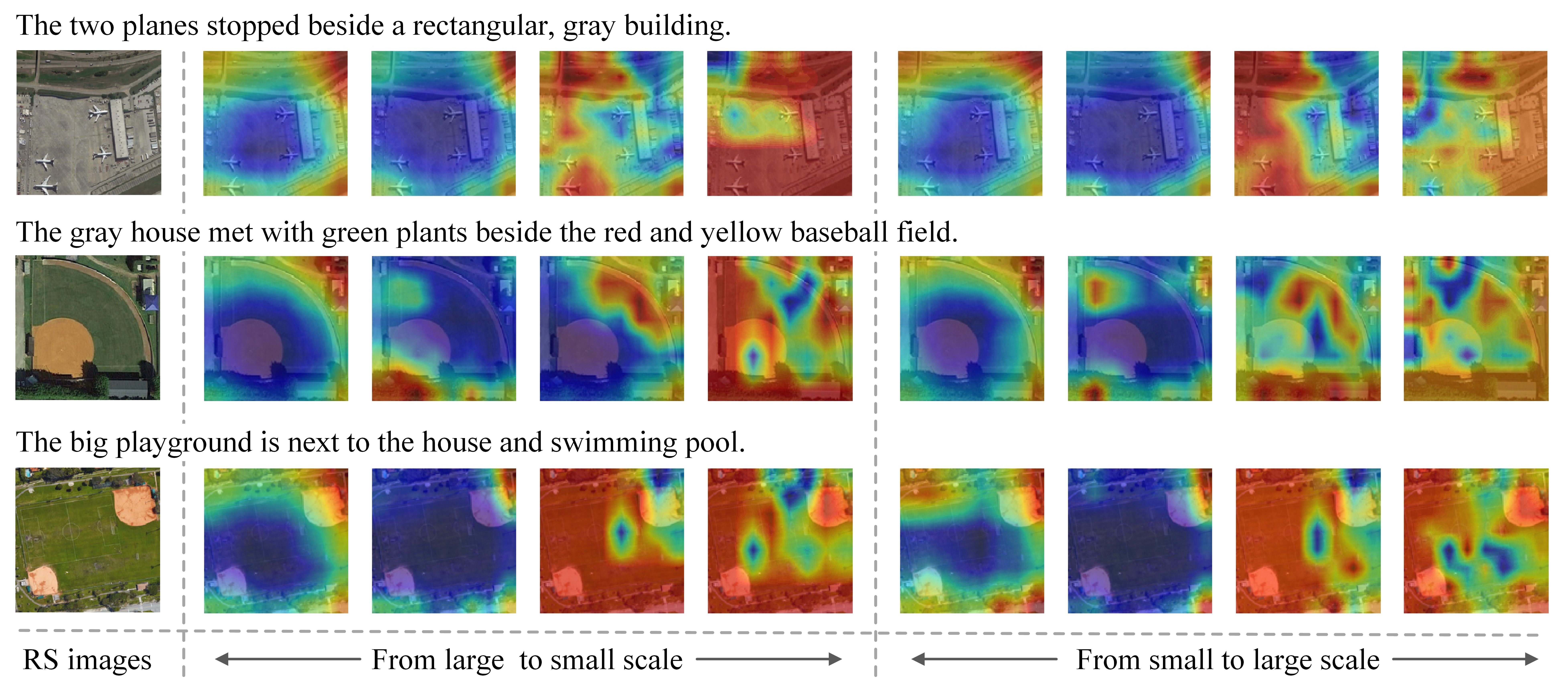}
	\caption{Visualization of salient mask output by the MBD in three RS images, which include three parts: The original RS images, the attention maps from large to small scales and the attention maps from small to large scales.}
	\label{fig3}
\end{figure*}
\begin{figure}
	\centering
	\includegraphics[width=0.55\linewidth]{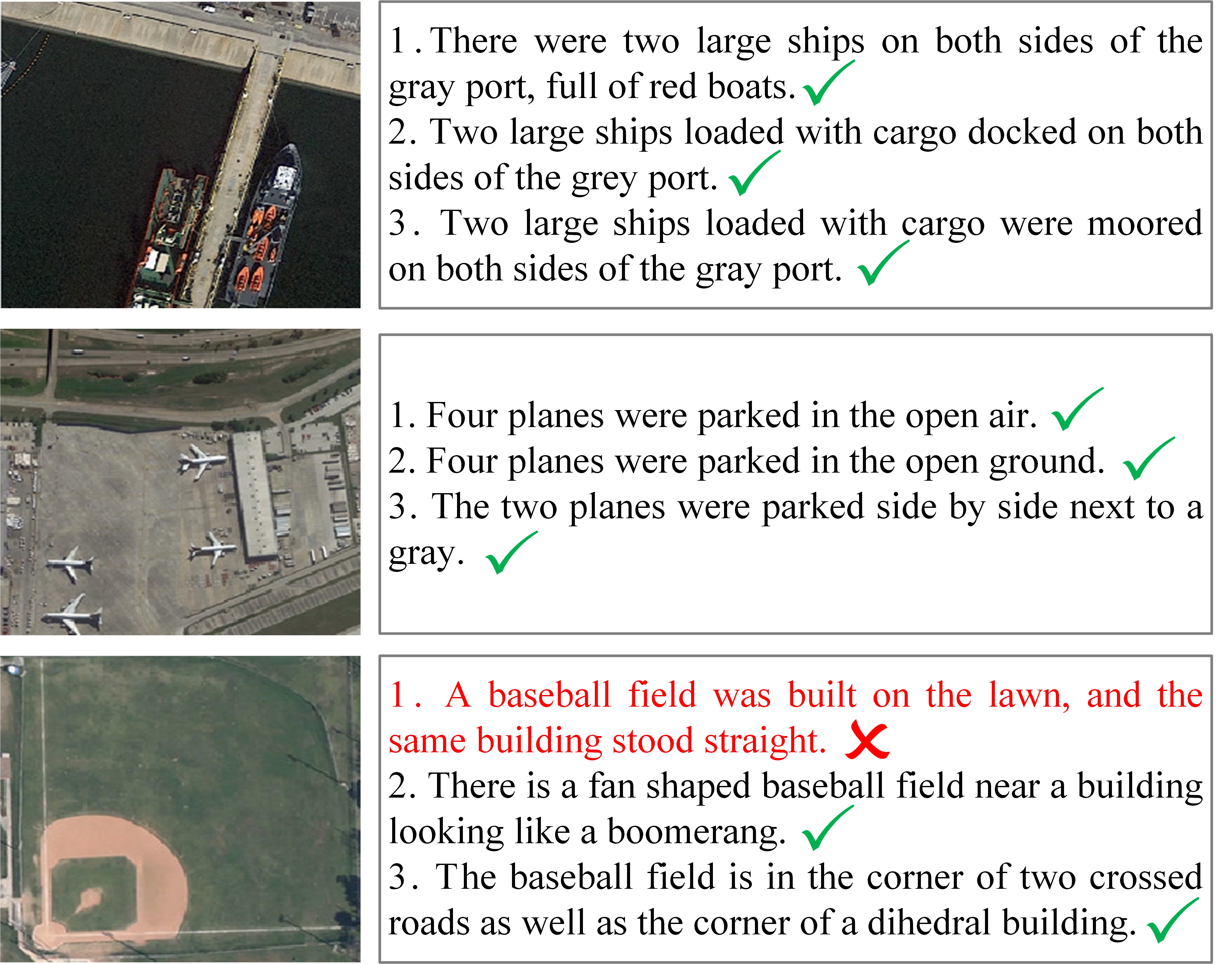}
	\caption{Top-3 image-to-text retrieval results on RSITMD dataset. The ground-truth texts are marked with green checks, and the wrong results are indicated by cross marks.}
	\label{fig3}
\end{figure}
\begin{figure}
	\centering
	\includegraphics[width=0.55\linewidth]{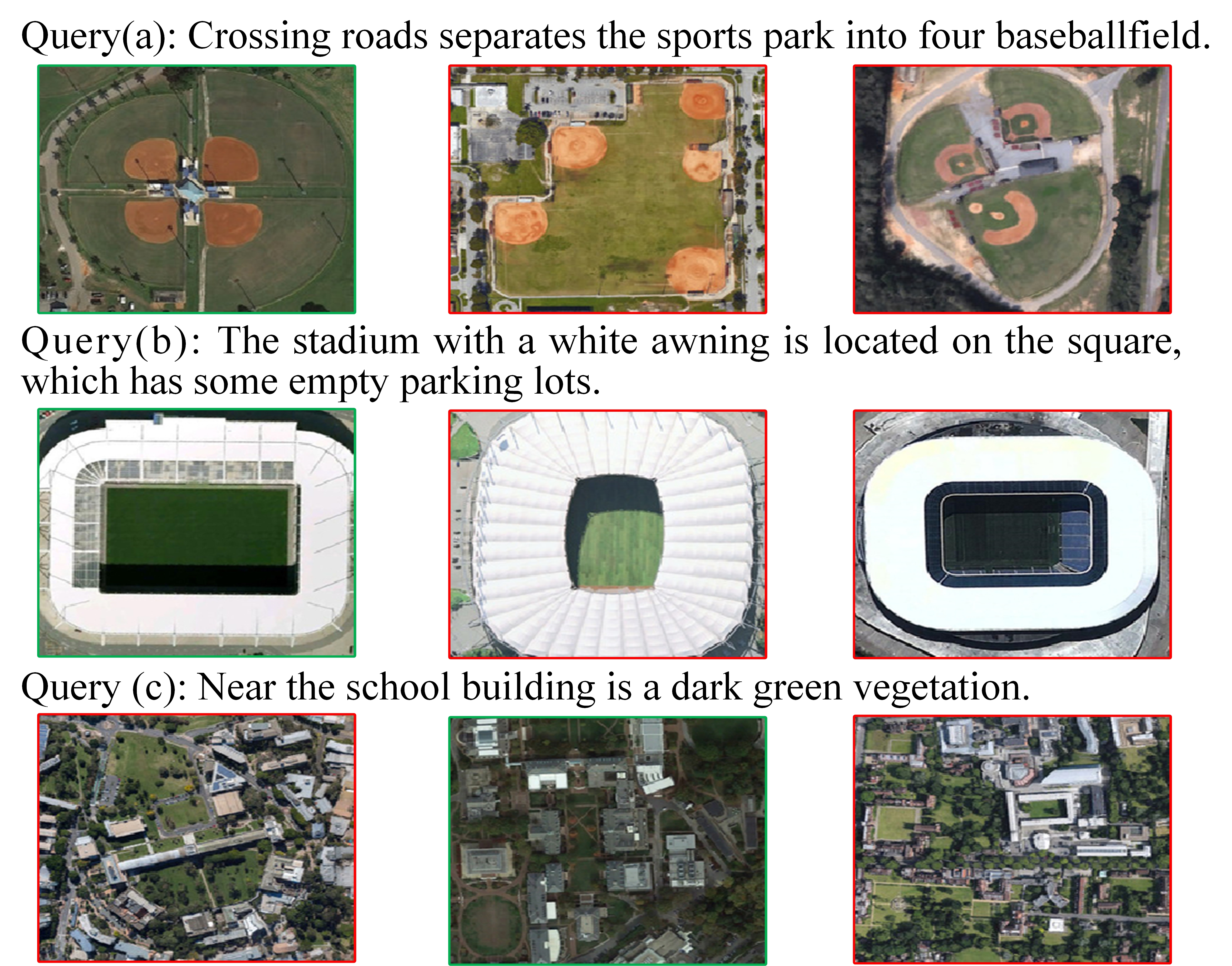}
	\caption{Top-3 text-to-image retrieval results on RSITMD dataset. The matched images are annotated in green boxes, and the false ones are in red.}
	\label{fig3}
\end{figure}
Compared with our model, the performance of \textbf{w/o STL} degrades dramatically. Particularly, It drops absolutely by 2.1 and 1.7 on R@10 of I2T and T2I for SSJDN, respectively, demonstrating the significance of the semantic-guided triple loss. Besides, our model achieves better results than \textbf{w/o LSD}, revealing that the label-supervised semantic decoupling module can enhance the representations by mining semantic-related semantic information and boost the model performance. Moreover, the performance drop of \textbf{w/o BSD} can be observed, indicating the vital importance of the bidirectional scale decoupling module as it can capture multi-scale discriminative clues. The \textbf{w/o all} method does not consider multi-scale distinguishing information and category supervision, resulting in inferior accuracy.
In general, our proposed model largely exceeds all variants on I2T and T2I retrieval, verifying the effectiveness and complementarity of four modules.

\subsubsection{Parameter Analysis}
The hyperparameter $\varepsilon$ is used to adjust the loss function, thereby increasing the retrieval probability of the same category data.  We vary $\varepsilon$ from 0.8 to 1.6 with intervals of 0.2 to search the value of $\varepsilon$ that optimizes the performance of the network. From Figure 4, we can observe that the model performs worst when the parameter is \textbf{0.8} because the category information has an adverse effect on the model training. If the value of the parameter is \textbf{1}, it means that the parameter has no action on the network. Moreover, the performance of the model reaches a saturation point when the value is \textbf{1.2}, and then begins to decline slightly.
Based on the above experiments, we choose $\varepsilon$ of 1.2 as the constant to increase the accuracy of retrieval by integrating category information.

\subsubsection{Comparison with Different Attention Mechanism}
In this section, we design an experiment to evaluate the performance of different attention approaches, as shown in Table 4. 
Especially, we select four different methods for combining attention: 1)\textbf{USD(L2S)}, noting the utilization of unidirectional (from large to small) scale decoupling module; 2)\textbf{USD(S2L)}, indicating the utilization of unidirectional (from small to large) scale decoupling module; 3)\textbf{MA}, representing that the attention maps are computed on image features at each scale; 4)\textbf{w/o MA}, standing for no attention method is employed. 

The comprehensive performance of the \textbf{USD(L2S)} and \textbf{USD(S2L)} decreases compared with our method, which indicates that it is necessary to perform bidirectional scale decoupling to explore distinguishing features from different perspectives. Although the \textbf{MA} conducts attention maps to probe salient features in RS images, the results are not satisfactory due to the large amount of redundancy contained in the attention maps. There is no doubt that the metrics $R@1$, $R@5$ and $R@1$ of \textbf{w/o MA} are the lowest in I2T and T2I retrieval because of the lack of mining of salient features.
The BSD module can eliminate redundant noise generated by multi-scale feature interaction by utilizing SFE and SGS units and yield encouraging retrieval performance.

\subsubsection{Performance of Different Fusion Method}
To better guide image and text feature training by category features, we conducted the following three experiments. 1)\textbf{Add}, referring to the element-wise addition of image (or text) features and category features; 2)\textbf{Concat}, indicating the concatenation; 3)\textbf{Multi}, representing the multiplication;

The experimental results are shown in Figure 5, and we can see that \textbf{Multi} outperforms the other two fusion method on the metrics $R@1$, $R@5$, and $R@10$ in I2T and T2I retrieval. It is concluded through analysis that multiplication further enhances the supervision of category features over image or text features, while \textbf{Add} and \textbf{Concat} simply combine the two features, resulting in inferior model performance than multiplication.

\subsection{Saliency Mask Visualization}
In this section, we visualize the attention map to analyze the function of the BSD module. In Figure 6, the first column is the original RS image, and columns 2 to 5 are attention maps generated by USD from large to small scale, similarly, columns 6 to 9 are the attention maps produced by the small to large scale USD module.

It can be observed from Figure 6 that the BSD module is inclined to focus on the effective region, as shown in the second column of the second image, the attention map concentrates on the "grey house" in the given query sentence. Besides, the attention map in each USD module emphasizes different salient regions by implementing decoupling operations, thereby mining differentiated effective cues. Moreover, there is also a discrepancy in the size of the attention regions of the USD module in two different directions, such as the upper left corner of the attention map in the second column and the sixth column in the third image.
It can be seen from these attention maps that the BSD module can filter redundant information generated in multi-scale interaction, improving the performance remarkably of RS image and text retrieval.
\subsection{Qualitative Results}
To qualitatively verify the effectiveness of SSJDN, several typical examples are illustrated from RSITMD in Figure 7 and Figure 8, respectively. For image-to-text retrieval, it can be observed that our model holds the ability to understand abstract short sentences or complex long sentences accurately. For text-to-image retrieval, the proposed model is robust to simple or complex images, which is mainly attributed to BSD, LSD and STL modules to fully exploit reliable and effective information. There are still some erroneous retrieval results, such as the rank-3 sentences of the third image query in Figure 8 and the rank-1 image of the Query (c) in Figure 9. The analysis concluded that these mistakes are caused by the high similarity between remote sensing data. In fact, it is even difficult for humans to distinguish the results, which is also the reason for the generally low accuracy for RS retrieval.

\section{Conclusion}

We have presented a novel SSJDN for cross-modal RS image retrieval. The main contribution of our method is that we propose a BSD module to adaptively extract potential features and suppress cumbersome features at other scales in a bidirectional pattern to exploit distinct clues. Besides, we utilize category semantic labels as the prior knowledge to generate category features, which are then combined with the original retrieval network to probe more effective class-related information. Finally, we design a semantic-guided triple loss, which tends to match the same class cross-modal data. 
The qualitative and quantitative results on popular four datasets show that our model achieves state-of-the-art performance in cross-modal retrieval tasks. 
In particular, in the ablation studies, the evaluation metrics mR of the SSJDN drop by 3.8, 4.9, and 1.8 without BSD, LSD, and STL operations, respectively, illustrating the effectiveness and reliability of the proposed method. 

We have verified the effectiveness of the category prior knowledge, and in the future, we plan to build a multi-category label to dynamically explore discriminative information in remote sensing images and texts, making it more interpretable and applicable.

\begin{acks}
This work was supported in part by the National Natural Science Foundation of China (62172376), the National Natural Science Foundation of China (62072418), the Fundamental Research Funds for the Central Universities (202042008) the National Key Research and Development Program of China (2021YFF0704000).
\end{acks}

\bibliographystyle{ACM-Reference-Format}
\bibliography{sample-base1}

\appendix

\end{document}